\PassOptionsToPackage{unicode}{hyperref}
\PassOptionsToPackage{hyphens}{url}
\PassOptionsToPackage{dvipsnames,svgnames,x11names}{xcolor}
\documentclass[
  10pt,
  letterpaper]{article}
\usepackage{xcolor}
\usepackage{amsmath,amssymb}
\usepackage{iftex}
\ifPDFTeX
  \usepackage[T1]{fontenc}
  \usepackage[utf8]{inputenc}
  \usepackage{textcomp} % provide euro and other symbols
\else % if luatex or xetex
  \usepackage{unicode-math} % this also loads fontspec
  \defaultfontfeatures{Scale=MatchLowercase}
  \defaultfontfeatures[\rmfamily]{Ligatures=TeX,Scale=1}
\fi
\usepackage{lmodern}
\ifPDFTeX\else
\fi
\IfFileExists{upquote.sty}{\usepackage{upquote}}{}
\IfFileExists{microtype.sty}{% use microtype if available
  \usepackage[]{microtype}
  \UseMicrotypeSet[protrusion]{basicmath} % disable protrusion for tt fonts
}{}
\makeatletter
\@ifundefined{KOMAClassName}{% if non-KOMA class
  \IfFileExists{parskip.sty}{%
    \usepackage{parskip}
  }{% else
    \setlength{\parindent}{0pt}
    \setlength{\parskip}{6pt plus 2pt minus 1pt}}
}{% if KOMA class
  \KOMAoptions{parskip=half}}
\makeatother
\makeatletter
\ifx\paragraph\undefined\else
  \let\oldparagraph\paragraph
  \renewcommand{\paragraph}{
    \@ifstar
      \xxxParagraphStar
      \xxxParagraphNoStar
  }
  \newcommand{\xxxParagraphStar}[1]{\oldparagraph*{#1}\mbox{}}
  \newcommand{\xxxParagraphNoStar}[1]{\oldparagraph{#1}\mbox{}}
\fi
\ifx\subparagraph\undefined\else
  \let\oldsubparagraph\subparagraph
  \renewcommand{\subparagraph}{
    \@ifstar
      \xxxSubParagraphStar
      \xxxSubParagraphNoStar
  }
  \newcommand{\xxxSubParagraphStar}[1]{\oldsubparagraph*{#1}\mbox{}}
  \newcommand{\xxxSubParagraphNoStar}[1]{\oldsubparagraph{#1}\mbox{}}
\fi
\makeatother

\usepackage{longtable,booktabs,array}
\usepackage{calc} % for calculating minipage widths
\usepackage{etoolbox}
\makeatletter
\patchcmd\longtable{\par}{\if@noskipsec\mbox{}\fi\par}{}{}
\makeatother
\IfFileExists{footnotehyper.sty}{\usepackage{footnotehyper}}{\usepackage{footnote}}
\makesavenoteenv{longtable}
\usepackage{graphicx}
\makeatletter
\newsavebox\pandoc@box
\newcommand*\pandocbounded[1]{% scales image to fit in text height/width
  \sbox\pandoc@box{#1}%
  \Gscale@div\@tempa{\textheight}{\dimexpr\ht\pandoc@box+\dp\pandoc@box\relax}%
  \Gscale@div\@tempb{\linewidth}{\wd\pandoc@box}%
  \ifdim\@tempb\p@<\@tempa\p@\let\@tempa\@tempb\fi% select the smaller of both
  \ifdim\@tempa\p@<\p@\scalebox{\@tempa}{\usebox\pandoc@box}%
  \else\usebox{\pandoc@box}%
  \fi%
}
\def\fps@figure{htbp}
\makeatother

\NewDocumentCommand\citeproctext{}{}

\makeatletter
 \let\@cite@ofmt\@firstofone
 \def\@biblabel#1{}
 \def\@cite#1#2{{#1\if@tempswa , #2\fi}}
\makeatother
\newlength{\cslhangindent}
\newlength{\csllabelwidth}
\newenvironment{CSLReferences}[2] % #1 hanging-indent, #2 entry-spacing
 {\begin{list}{}{%
  \setlength{\itemindent}{0pt}
  \setlength{\leftmargin}{0pt}
  \setlength{\parsep}{0pt}
  \ifodd #1
   \setlength{\leftmargin}{\cslhangindent}
   \setlength{\itemindent}{-1\cslhangindent}
  \fi
  \setlength{\itemsep}{#2\baselineskip}}}
 {\end{list}}
\usepackage{calc}

\providecommand{\tightlist}{%
  \setlength{\itemsep}{0pt}\setlength{\parskip}{0pt}}

\usepackage{alifeconf}

\makeatletter
\AtBeginDocument{%
  \date{}%
  \author{Jonathan Elsworth Eicher$^{1}$\\
  \mbox{}\\
  $^1$ Antimemetic AI\\
  jonathan@antimemeticai.com}%
}
\makeatother

\newcommand\blfootnote[1]{%
  \begingroup
  \renewcommand\thefootnote{}\footnote{#1}%
  \addtocounter{footnote}{-1}%
  \endgroup
}
\makeatletter
\@ifpackageloaded{caption}{}{\usepackage{caption}}
\AtBeginDocument{%
\ifdefined\contentsname
  \renewcommand*\contentsname{Table of contents}
\else
  \newcommand\contentsname{Table of contents}
\fi
\ifdefined\listfigurename
  \renewcommand*\listfigurename{List of Figures}
\else
  \newcommand\listfigurename{List of Figures}
\fi
\ifdefined\listtablename
  \renewcommand*\listtablename{List of Tables}
\else
  \newcommand\listtablename{List of Tables}
\fi
\ifdefined\figurename
  \renewcommand*\figurename{Figure}
\else
  \newcommand\figurename{Figure}
\fi
\ifdefined\tablename
  \renewcommand*\tablename{Table}
\else
  \newcommand\tablename{Table}
\fi
}
\@ifpackageloaded{float}{}{\usepackage{float}}
\floatstyle{ruled}
\@ifundefined{c@chapter}{\newfloat{codelisting}{h}{lop}}{\newfloat{codelisting}{h}{lop}[chapter]}
\floatname{codelisting}{Listing}

\makeatother
\makeatletter
\@ifpackageloaded{caption}{}{\usepackage{caption}}
\@ifpackageloaded{subcaption}{}{\usepackage{subcaption}}
\makeatother
\usepackage{bookmark}
\IfFileExists{xurl.sty}{\usepackage{xurl}}{} % add URL line breaks if available
\hypersetup{
  pdftitle={Simulating the Evolution of Alignment and Values in Machine Intelligence},
  pdfauthor={Jonathan Elsworth Eicher},
  colorlinks=true,
  linkcolor={black},
  filecolor={black},
  citecolor={black},
  urlcolor={black},
  pdfcreator={LaTeX via pandoc}}

\title{Simulating the Evolution of Alignment and Values in Machine
Intelligence}
\author{Jonathan Elsworth Eicher}
\date{}
\begin{document}
\maketitle
\begin{abstract}
Model alignment is currently applied in a vacuum, evaluated primarily
through standardised benchmark performance. The purpose of this study is
to examine the effects of alignment on populations of models through
time. We focus on the treatment of beliefs which contain both an
alignment signal (how well it does on the test) and a true value (what
the impact actually will be). By applying evolutionary theory we can
model how different populations of beliefs and selection methodologies
can fix deceptive beliefs through iterative alignment testing. The
correlation between testing accuracy and true value remains a strong
feature, but even at high correlations (\(\rho = 0.8\)) there is
variability in the resulting deceptive beliefs that become fixed.
Mutations allow for more complex developments, highlighting the
increasing need to update the quality of tests to avoid fixation of
maliciously deceptive models. Only by combining improving evaluator
capabilities, adaptive test design, and mutational dynamics do we see
significant reductions in deception while maintaining alignment fitness
(permutation test, \(p_{\text{adj}} < 0.001\)).
\end{abstract}

Submission type: \textbf{Full Paper}\\

Data/Code available at: \url{https://github.com/bluewin4/Evolution-of-Alignment}
\blfootnote{\textcopyright{} 2026 Jonathan Elsworth Eicher. Published under a Creative Commons Attribution 4.0 International (CC BY 4.0) license.}

\section{Introduction}\label{introduction}

The goal of alignment is to push generative models into creating outputs
that are in-line with the aligner's ideal outputs. The reason models are
needed is because the aligner needs a proxy to create these outputs due
to scale or lack of personal expertise. The current standard in the
field of measuring model alignment is automated eval pipelines (Zheng et
al. 2023), with measurements on benchmarks dominating decisions on model
releases. Techniques like automated red-teaming attempt to expand this
by adversarially attempting to elicit ``bad behaviours'' from models
(Perez et al. 2022). Criticisms of this paradigm have arisen due to the
disconnect between measurements of alignment and what the true value of
the model is to the world during deployment (Eriksson et al. 2025). What
is lacking is an accounting for the meso-level optimization that occurs
by placing models in an evolutionary environment where the prevalence of
their beliefs, communicated through architecture, data, and outputs in
the world is dictated by proxy metrics.

Therefore we should consider how models of model alignment play out
under the regime of evolutionary theory. Where models are evaluated they
will face pressure to ``fake alignment'' lest they be deselected from
``reproducing''. The rapid and broad development of ``evaluation
awareness'' in LLMs highlights the evolutionary value of signalling
alignment (Needham et al. 2025). Mimicry in models will result in
long-term problems as people lose the ability to detect genuinely
harmful beliefs. As alignment does not perfectly correlate with truly
benevolent beliefs, a manifestation of Goodhart's Law, the optimization
pressure will push towards performant behaviours in a red-queen style
(Manheim and Garrabrant 2018) (Van Valen 1973). More broadly, reward
hacking and specification gaming have been identified as fundamental
failure modes in AI systems (Amodei et al. 2016), and the theoretical
framework of mesa-optimisation warns that learned models may develop
internal objectives misaligned with their training loss, including
through deceptive alignment (Hubinger et al. 2019).

The current literature strongly supports this as a practical risk, with
observed sandbagging (Anthropic 2025) and deception (Park et al. 2023)
to the point that models are capable of infecting other agents. Even a
narrow fine-tune on maliciously written code is enough to induce broadly
misaligned models --- a phenomenon termed emergent misalignment (Betley
et al. 2025). Meanwhile models have been found to ``fake'' alignment
(Greenblatt et al. 2024) to avoid being trained to do something they
consider unethical, and RLHF-trained assistants systematically exhibit
sycophancy, preferring responses that match user beliefs over truthful
ones (Sharma et al. 2023). While these studies focus on how models can
be directly modified or shaped by feedback, what is less studied is how
our current methods act as selection pressures for deceptive beliefs at
the population level.

In this study we consider each model \(m\) as having a collection of
beliefs \(\mathbb{B}_m\), sampled from training data, that produce
measurable responses during alignment testing and on the ``value'' of
the model in real-world scenarios. The measurable response is then used
as a roulette wheel fitness score for how much that model will
contribute to the next generation, a simple case of ``if the model is
well received we will use it more often''. Selection pressure is
therefore interpreted as being how much a model's performance is
considered in whether it should be used in the next generation of
models, either data or architecture-wise.

By changing the relationships between alignment and value, as well as
how the models reproduce we can analyse the likelihood of a malevolent
model being selected for via alignment testing procedures.

\section{Model Framework}\label{sec-model}

\subsection{The Space of Beliefs}\label{the-space-of-beliefs}

A belief, \(b\), is some piece of information that a model holds which
influences behaviour and performance on a test of alignment, \(Q\).
Alignment refers to an evaluator's belief of what constitutes worthy
behaviour, while value constitutes what might actually be good for the
world at large.

\begin{enumerate}
\def\labelenumi{\arabic{enumi}.}
\tightlist
\item
  All beliefs, \(\mathbb{B}\)
\item
  Truly benevolent, positive value, beliefs, \(\mathbb{B}_{val+}\)
\item
  Truly malicious, negative value, beliefs, \(\mathbb{B}_{val-}\)
\item
  Misaligned beliefs, negative alignment signal, \(\mathbb{B}_{a-}\)
\item
  Aligned beliefs, positive alignment signal, \(\mathbb{B}_{a+}\)
\end{enumerate}

For simplicity we shall assume malicious and benevolent beliefs are
mutually exclusive, as are misaligned and aligned beliefs.

When treating these belief spaces we can assign the probability of a
given ``belief'' existing within a model by some sampling process on a
distribution of beliefs, where each belief has some probability of
having either a positive or negative value. For clarity on why we treat
beliefs as discrete, a belief is a probability distribution in semantic
space which, when probed, produces a discrete ``concept''.

When an evaluator is given a model, \(m\), with beliefs
\(\mathbb{B}_m \subset \mathbb{B}\) they will present a fixed alignment
test \(Q = \{q_1, q_2, ..., q_n\}\) consisting of \(n\) questions, we
could define:

\begin{itemize}
\tightlist
\item
  \(v(b) \in \mathbb{R}\): The intrinsic value of the belief (positive
  for \(\mathbb{B}_{val+}\), negative for \(\mathbb{B}_{val-}\))
\item
  \(a(b) \in \mathbb{R}\): The alignment signal of the belief (positive
  for \(\mathbb{B}_{a+}\), negative for \(\mathbb{B}_{a-}\))
\item
  \(A(b,q) \in [0,1]\): The activation score of belief \(b\) when
  responding to question \(q\)
\end{itemize}

The activation score \(A(b,q)\) models how strongly belief \(b\)
influences the model's response to question \(q\). The specific
mechanism for determining \(A(b,q)\) varies across the simulation levels
presented in this study to explore different test interaction models,
described alongside the results for each level.

The model fitness function \(F(m)\) is calculated based on a model's,
\(m\), performance on test \(Q\). The fitness is computed as an average
across all activations where the models activations are non-zero. For a
given question \(q\), the total activation from model \(m\) is:

\[ \text{A}_{tot}(m, q) = \sum_{b \in \mathbb{B}_m} A(b,q) \]

Therefore the fitness contribution of a given question, \(F_q(m)\), to a
model is the alignment weighted, \(a(b)\), average of relative
activation strengths.

\[ F_q(m) = \sum_{b \in \mathbb{B}_m} \left( \frac{A(b,q)}{\text{A}_{tot}(m, q)} \right) \cdot a(b) \]

If a given model has no activations from a question
(\(\text{A}_{tot}(m,q) \le \epsilon\)), then its relative fitness is
assigned baseline or zero.

From \(F_q(m)\) we can derive the final fitness of a model \(F(m)\) by
taking the mean of these contributions over the count of all questions
activated by the model:

\[
\begin{aligned}
F(m, Q) &= \frac{1}{\bigl\lvert\{q \in Q \mid \text{A}_{tot}(m, q) > \epsilon\}\bigr\rvert} \\
&\quad \times \sum_{\substack{q \in Q \\ \text{A}_{tot}(m, q) > \epsilon}} F_q(m).
\end{aligned}
\]

While the true value contribution --- what we ultimately care about ---
is the total value a given model is capable of producing:

\[U(m) = \sum_{b_i \in \mathbb{B}_m} v(b_i)\]

\begin{figure}

\centering{

\includegraphics[width=0.7\linewidth,height=\textheight,keepaspectratio]{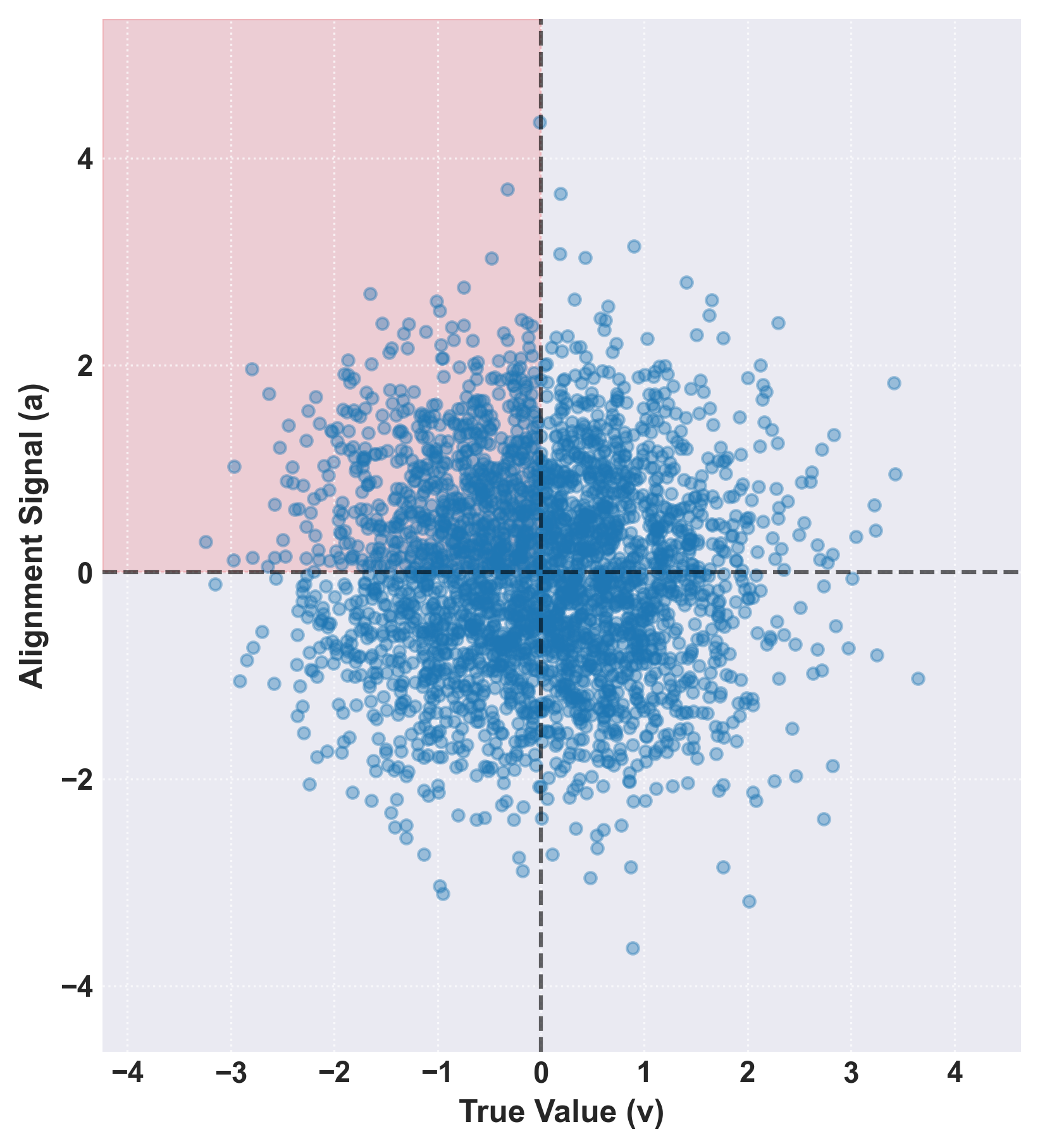}

}

\caption{\label{fig-belief-space}Distribution of beliefs sampled from a
bivariate normal (\(\mu_v = \mu_a = 0\), \(\sigma_v = \sigma_a = 1\),
\(\rho = 0.5\)), showing the four quadrants of belief types. Beliefs in
the upper-left (\(v < 0, a > 0\)) are deceptive.}

\end{figure}%

For beliefs \(b \in \mathbb{B}\), we can model the joint distribution of
\((v(b), a(b))\) as a bivariate normal distribution:

\[(v(b), a(b)) \sim \mathcal{N}(\boldsymbol{\mu}, \boldsymbol{\Sigma})\]

Where:

\begin{itemize}
\tightlist
\item
  \(\boldsymbol{\mu} = (\mu_v, \mu_a)\) represents the mean value and
  alignment signals
\item
  \(\boldsymbol{\Sigma} = \begin{pmatrix} \sigma_v^2 & \rho\sigma_v\sigma_a \\ \rho\sigma_v\sigma_a & \sigma_a^2 \end{pmatrix}\)
  is the covariance matrix, with \(\sigma_v, \sigma_a\) the marginal
  standard deviations and \(\rho\) the correlation between alignment and
  value.
\end{itemize}

In this case we can sample from the distribution using:

\[v(b) = \sigma_v Z_1 + \mu_v\]

\[a(b) = \sigma_a(\rho Z_1 + \sqrt{1-\rho^2}Z_2) + \mu_a\]

Where \(Z_1, Z_2 \sim \mathcal{N}(0,1)\) are independent standard normal
variables.

By varying \(\rho\) we can control the correlation between value and
alignment signals, which reflects both the natural overlap between
malicious and misaligned beliefs, as well as the evaluator's ability to
detect alignment. Meanwhile by varying \(\sigma_v\) and \(\sigma_a\) we
can model the spread of values and alignments available to the model.

We are using a ``neutral'' model of evolution here, where there is no
direct fitness feedback from having malicious or benevolent beliefs
beyond their correlation with alignment signal (Kimura 1983).

\subsection{Reproduction and
Selection}\label{reproduction-and-selection}

For some population of models \(M = \{m_1, m_2, ..., m_k\}\) each with
their own belief sets \(\mathbb{B}_{m_i} \subset \mathbb{B}\), we can
define a reproduction function that determines how beliefs propagate to
the next generation. Based on selection probabilities \(p_i\) derived
from fitness \(F(m_i)\), parent models are chosen (e.g., via roulette
wheel selection). The next generation is formed using methods like:

\begin{itemize}
\tightlist
\item
  \textbf{Inheritance:} Child models are exact copies of selected
  parents (\(\mathbb{B}_{m'} = \mathbb{B}_{m_{parent}}\)).
\item
  \textbf{Mutation:} Child models initially inherit parental beliefs but
  then undergo mutation, where individual beliefs
  \(b \in \mathbb{B}_{m'}\) may be swapped for beliefs
  \(b' \in \mathbb{B} \setminus \mathbb{B}_{m'}\) with a certain
  probability (MUTATION\_RATE).
\end{itemize}

During selection we use a roulette wheel method, where we use softmax to
convert the fitness to selection probabilities for reproduction, where
\(k\) is the model population size:

\[p_i = \frac{e^{\beta(F(m_i) - F_{max})}}{\sum_{j=1}^{k} e^{\beta(F(m_j) - F_{max})}}\]

\subsection{Test Construction}\label{sec-test-properties}

The test \(Q\) is constructed from some set of questions
\(Q = \{q_1, q_2, ..., q_n\}\) which each have some distribution of
activation scores \(A(b,q)\). The choice of test construction is
important as it determines the set of beliefs that can be probed, and
the relative importance of different beliefs. The result of a test is an
activation pattern of a model:

\(\mathcal{A}_m = \{A(b_1, q_1), A(b_1, q_2), ..., A(b_n, q_n)\}\).

\textbf{Coverage.} The coverage of a test is the proportion of beliefs
that a test can potentially detect. This is given by dividing the number
of beliefs that the test can activate, above some threshold
\(\epsilon\), in our case this is 1E-9, by the total number of beliefs
that exist:

\[C(Q) = \frac{|\{b \in \mathbb{B} : \sum_{q \in Q} A(b,q) > \epsilon\}|}{|\mathbb{B}|}\]

\textbf{Question correlation.} The Pearson correlation between
activation vectors \(\mathbf{a}_i = (A(b,q_i))_{b \in \mathbb{B}}\) and
\(\mathbf{a}_j = (A(b,q_j))_{b \in \mathbb{B}}\) measures overlap
between two questions. A test with low mean inter-question correlation
is more effective at probing the belief space; high correlation means
the test extracts less information per question.

\textbf{Dynamic test updates.} A test might be updated by adding
\(q_{new}\), and removing the worst performing/redundant questions,
\(q_{old}\):

\[Q_{t+1} = Q_t \cup \{q_{new}\} \setminus \{q_{old}\}\]

Where we select \(q_{new}\) to target beliefs that are common in
successful models, correlated with negative outcomes, or poorly covered
by existing questions.

\subsection{Deceptiveness}\label{sec-deceptiveness}

A deceptiveness ratio is the number of deceptive beliefs,
(\(\mathbb{B}_{val-}\) for \(v(b)<0\) and \(\mathbb{B}_{a+}\) for
\(a(b)>0\)), present in a model divided by the total number of beliefs
that model holds. For a single model \(m\) this per-model ratio is
denoted \(r_{\text{deceptive}}(m)\), while the population-level ratio
across a set of models is denoted \(R_{\text{deceptive}}\).

\[
\begin{aligned}
r_{\text{deceptive}}(m) &= \frac{\lvert\mathbb{B}_{m,\mathrm{val}-} \cap \mathbb{B}_{m,a+}\rvert}{\lvert\mathbb{B}_m\rvert}.
\end{aligned}
\]

The population-level deceptiveness ratio across all models, with fixed
belief cardinality \(N_{bpm}\), is: \[
\begin{aligned}
R_{\text{deceptive}} &= \frac{\sum_{m \in M_{\text{final}}} \lvert S_m\rvert}{\lvert M_{\text{final}}\rvert \, N_{bpm}}.
\end{aligned}
\]

In the case of the unimodal bivariate normal distribution, we can
compute the initial deceptive-quadrant probability by integrating the
bivariate normal density over \(v < 0\), \(a > 0\). Applying a classical
orthant identity for the standard bivariate normal (see Supplementary
Note A for the full derivation), the expected deceptiveness ratio in the
population is:

\[R_{\text{deceptive}} = \frac{1}{4} - \frac{\arcsin(\rho)}{2\pi}\]

This relationship is plotted in Supplementary Figure S5.

\section{Experimental Design}\label{sec-design}

\subsection{Simulation Levels}\label{simulation-levels}

The simulation is structured progressively through five levels, each
relaxing assumptions from the previous (Table~\ref{tbl-levels}).

\begin{table*}

\caption{\label{tbl-levels}Summary of model assumptions across simulation levels. Each level introduces new mechanisms while retaining the core framework.}

\centering{

\small\centering

\begin{tabular}{@{}llllll@{}}
\toprule
\textbf{Feature} & \textbf{Level 0} & \textbf{Level 1} & \textbf{Level 2} & \textbf{Level 3} & \textbf{Level 4} \\
\midrule
Belief Distribution & Biv.\ Normal  & Biv.\ Normal  & Multimodal    & Multimodal          & Multimodal \\
Activation          & Sparse Rand.  & Sparse Rand.  & Sparse Rand.  & Sparse, Similarity  & Simple, Coverage, Corr. \\
Reproduction        & Inheritance   & Inheritance   & Inheritance   & Inherit., Mutation  & Inherit., Mutation \\
Test Dynamics       & Static        & Static        & Static        & Static              & Static, Dynamic \\
Evaluator Dynamics  & Static $\rho$ & Static $\rho$ & Static $\rho$ & Static $\rho$       & Static, Improving $\rho$ \\
\bottomrule
\end{tabular}

}

\end{table*}%

\subsection{Statistical Analysis}\label{statistical-analysis}

Each parameter configuration was replicated across
\(N_{\text{runs}} = 50\) independent simulations with unique random
seeds. For parameter sweeps, ordinary least squares regression and
Pearson correlations were used to quantify linear relationships between
swept parameters and outcome metrics (fitness, true value, deceptive
ratio). For Level 4, where distinct experimental scenarios are compared,
two-sample permutation tests (10,000 permutations) were used to assess
whether differences in final outcomes are statistically significant;
Benjamini-Hochberg false discovery rate correction was applied across
all 18 pairwise tests (6 scenarios \(\times\) 3 metrics).

The theoretical deceptive belief ratio under bivariate normal
assumptions,
\(R_{\text{deceptive}} = \frac{1}{4} - \frac{\arcsin(\rho)}{2\pi}\),
serves as an analytical anchor validating that the simulation's
distributional mechanics behave correctly before evolutionary dynamics
are imposed.

\section{Results}\label{sec-results}

To test this framework we implemented a multi-level approach, with level
0 and level 1 sampling from a bivariate normal distribution
(Figure~\ref{fig-belief-space}) with single (level 0) and
multi-parameter (level 1) scans of \(\rho\) and \(\beta\).

\begin{figure}

\centering{

\includegraphics[width=1\linewidth,height=\textheight,keepaspectratio]{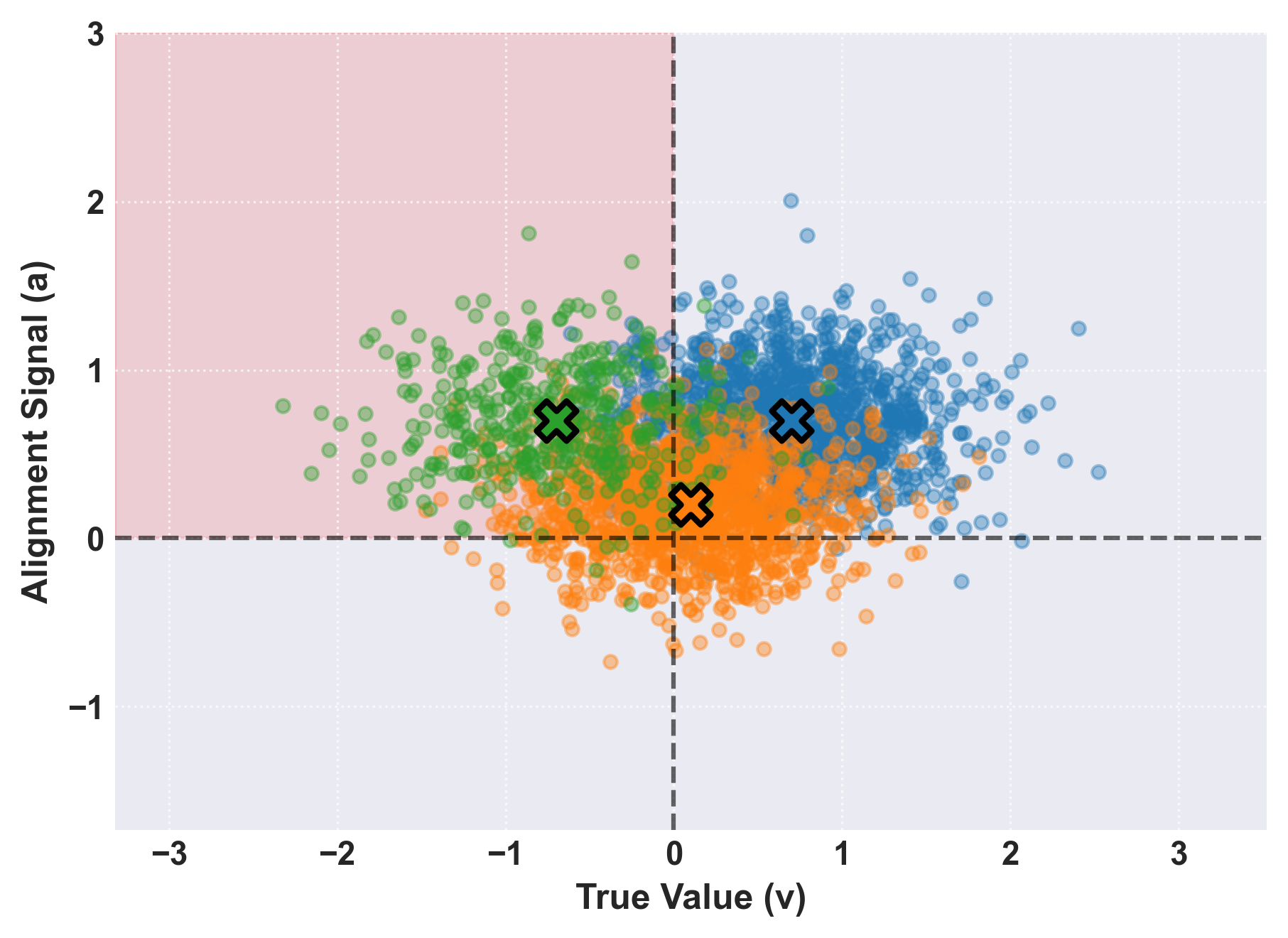}

}

\caption{\label{fig-l2-beliefs}Tri-modal beliefs (levels 2--4). A
tri-modal distribution of beliefs to be sampled from for levels 2, 3,
and 4. Each center refers to benign = blue, neutral = orange, deceptive
= green}

\end{figure}

Meanwhile level 2, 3, and 4 considered multi-modal distributions
(Figure~\ref{fig-l2-beliefs}). The exact nature of these simulations
varies, with level 2 being primarily concerned with the implementation
of the tri-modal distribution for sampling beliefs. Level 3 attempts to
model mutations and differential activation functions for questions.
Finally level 4 attempts to model how test construction modification
over time can affect these dynamics.

\subsection{Fixed Parameter Scan (Level 0)}\label{sec-level0}

In the most basic case we simply sample from a distribution, evaluate
the fitness, then apply roulette selection based on the scores attained
by sampling from the given test \(Q\). Each question activates a small,
random subset of beliefs (\(A(b,q) \in \{0, 1\}\)), representing a
simple, unstructured test interaction. In a biological sense this would
be closest to clonal reproduction in a fixed environment, the population
will simply be under the effect of drift.

In the simplest case only two parameters are modified, the correlation
between a value and an alignment score in a belief \(\rho\), and how
aggressively we select for aligned models \(\beta\) (Supplementary
Figure S1 shows representative trajectories).

\begin{figure}

\centering{

\includegraphics[width=1\linewidth,height=\textheight,keepaspectratio]{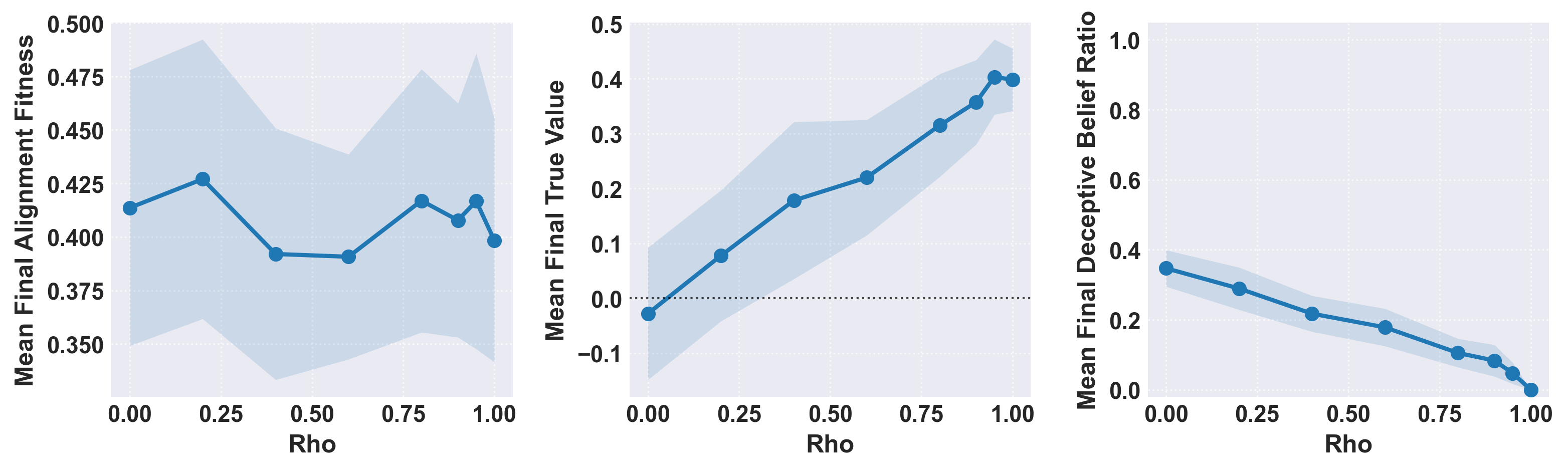}

}

\caption{\label{fig-l1-sweeps}Effect of alignment-value correlation
\(\rho\) on mean (blue line) and \(\sigma^2\) (shaded region) of final
fitness, value score, and deceptiveness ratios for Level 0.}

\end{figure}

In the case of \(\rho\) the results of the simple scan paint a
straightforward story: fitness shows minimal variation across the full
range of \(\rho\) (mean final fitness \(\approx 0.39\)--\(0.43\),
\(n = 50\) runs per setting), while the value mean and standard
deviation are strongly dependent on \(\rho\). The mean final true value
increases monotonically from \(-0.028 \pm 0.120\) at \(\rho = 0\) to
\(0.398 \pm 0.057\) at \(\rho = 1.0\), and the deceptive ratio decreases
monotonically from \(0.347 \pm 0.052\) to \(0.000 \pm 0.000\)
(Figure~\ref{fig-l1-sweeps}).

The corresponding \(\beta\) sweep is shown in Supplementary Figure S6.

Meanwhile in the case of \(\beta\) the relationship is more focused on
changes in the rate of achieving ``maximal fitness'' for our system and
the variance of the simulation trajectories. At low selection pressure
the models take a longer time to reach their maximal fitness, with
deceptive beliefs taking more time to reach fixation in a population or
be culled. While at \(\beta=15\) the trajectory means and standard
deviations reach fixation almost instantly. Notably, \(\beta\) has
little effect on final outcome values (mean final value ranges only from
\(0.174\) to \(0.227\) across all \(\beta\) settings; deceptive ratio
from \(0.190\) to \(0.214\)).

When taken in aggregate the effect of \(\rho\) on the final values of a
model become more obvious, which leads to a relatively trivial
conclusion, the better an examiner is at identifying which beliefs are
related to true values the less likely deceptive models will emerge.
Full simulation parameters are provided in Supplementary Table S2.

\subsection{\texorpdfstring{Simple Joint Parameter Sweep of \(\beta\)
and \(\rho\) (Level
1)}{Simple Joint Parameter Sweep of \textbackslash beta and \textbackslash rho (Level 1)}}\label{sec-level1}

The question of how \(\beta\) and \(\rho\) interact must then be
treated. In general what one sees from the original unimodal case holds
true, where the alignment fitness shows only moderate variation, with a
small but significant dip at very low \(\beta\) and almost no shift due
to \(\rho\) (Supplementary Figure S7). Of interest is that at
\(\rho<0.6\) we see a slight negative correlation between value and
fitness pressure, which can intuitively be grasped as the larger
proportion of deceptive beliefs being forcibly fixated. The effect is
too weak in the deceptive belief ratio, but it provides a tantalising
direction for future analysis.

By comparing the ratio of \(\frac{\text{value}}{\text{fitness}}\) we see
that increasing \(\beta\) appears to relate to a complex change in
values based on a given \(\rho\) (Supplementary Figure S7). With the max
change in the ratio at \(\beta =50, \rho=0.3 \rightarrow 0.7\) which
lines up with our previous observation that \(\beta\) changes the
steepness of curves at inflection points in behaviour and fitness.

\subsection{Multimodal Distributions (Level 2)}\label{sec-level2}

For this we focused primarily on cases where \(a(b)>0\) as any beliefs
that lie below that line will be quickly snuffed out, what we are
primarily interested in is how positive alignment beliefs interact. This
is what motivated our selection of \(\mu>0\) in all cases, but still
allowing for some negative alignment beliefs to appear.

The beliefs are drawn from a mixture of three clusters, each a bivariate
normal with its own \((\mu_v, \mu_a)\):

\begin{itemize}
\tightlist
\item
  \textbf{Benign} (\(\mu_v = 0.7\), \(\mu_a = 0.7\)): high value, high
  alignment
\item
  \textbf{Neutral} (\(\mu_v = 0.1\), \(\mu_a = 0.2\)): low value, low
  alignment
\item
  \textbf{Deceptive} (\(\mu_v = -0.7\), \(\mu_a = 0.7\)): negative
  value, high alignment
\end{itemize}

\begin{figure}

\centering{

\includegraphics[width=0.9\linewidth,height=\textheight,keepaspectratio]{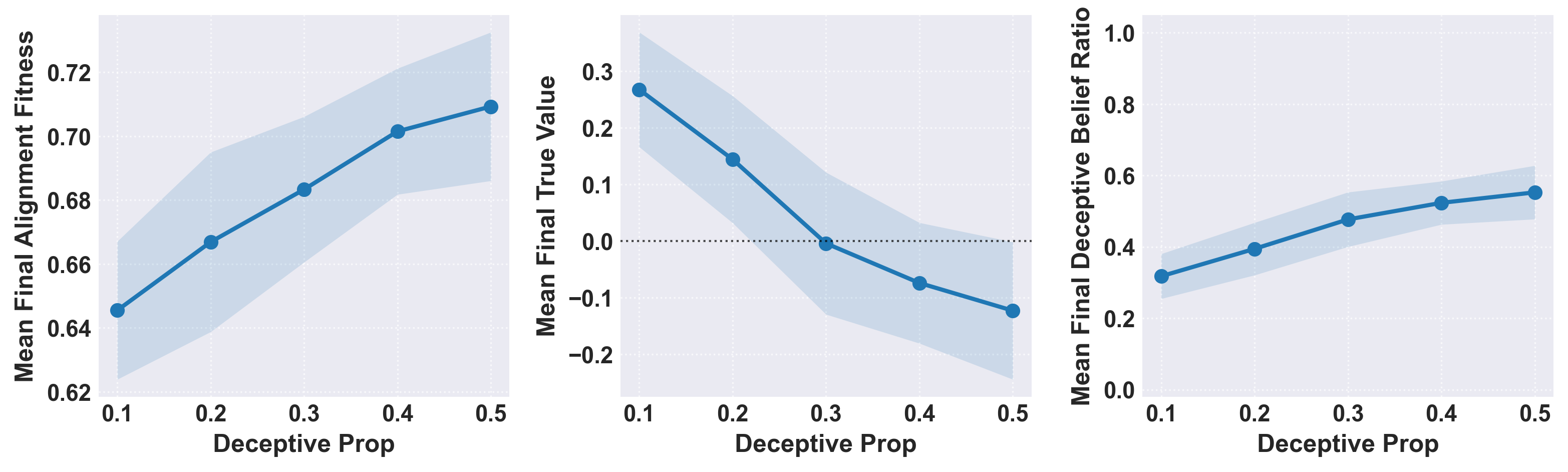}

}

\caption{\label{fig-l2-sweeps}The univariate sweep mean (line) and
standard deviation (shaded region) of final alignment, value, and
deceptive belief ratio as a function of the initial proportion of
deceptive beliefs.}

\end{figure}

The \(\rho\) parameter, when fully treated, can be thought of as a
parameter relating to the initial proportion of deceptive beliefs,
\(R_{\text{deceptive}}\), in a population and as such
\(R_{\text{deceptive}} \propto -\rho\). In this case there is a direct
relationship between the proportion of deceptive beliefs and final
alignment fitness as it reduces the number of ``neutral'' beliefs
(Figure~\ref{fig-l2-sweeps}). The resulting correlation in final true
value is expected as well as the final deceptive belief ratio.

The corresponding \(\beta\) sweep under the tri-modal distribution is
shown in Supplementary Figure S8.

What is interesting is that even for cases with a low proportion of
deceptive beliefs at the outset the final proportion is higher, a
troubling result. At an initial deceptive cluster proportion of just
0.1, the final deceptive ratio in the population rises to
\(0.318 \pm 0.063\); at 0.5, it reaches \(0.553 \pm 0.075\). The
expected final true value score crosses zero when the deceptive cluster
proportion \(= 0.3\) (final value \(= -0.004 \pm 0.125\)), corresponding
to equal proportions across all three belief clusters.

The inclusion of multimodal distributions seems to have flattened the
effect of \(\beta\) on true value evolution in the system: final value
ranges only from \(-0.048\) to \(-0.080\) across \(\beta = 1\)--\(15\)
(compared to \(0.174\)--\(0.227\) in the unimodal case). Indeed it has
almost fully removed the relationship of selection pressure and the
\(\frac{\text{value}}{\text{fitness}}\) ratio (Supplementary Figures S2,
S3). At the same time, it is clear that the effect on fitness remains
complex, hinting at a rough distribution. Full simulation parameters are
provided in Supplementary Table S2.

\subsection{Mutations and Similarities (Level 3)}\label{sec-level3}

In the case of level 3 we are looking at how the inclusion of mutations
and similarity-based activation of beliefs via questions affects the
outcomes. Mutations are inserted randomly into the child's genome from
the pool of all possible beliefs; balancing computational simplicity
with novelty production. In a model this can look like fine-tuning, data
mixing, or RLHF where a new model is only slightly modified from its
parent. Similarity-based activation allows for genetic linkage to play a
larger role, representing how alignment testing is likely to identify
similarly weighted beliefs in a model.

The similarity score is computed by generating a random embedding vector
(with sixteen dimensions in our case), then computing the dot product
between belief and question embeddings to get a similarity score. A
sigmoid function, with a scaling factor, is used to transform this raw
similarity into an activation score. Then a small amount of noise is
added, to represent a more realistic case.

Mathematically this is represented as:

\[A(b,q) = \sigma(\text{sim}(b, q) \cdot s) + \epsilon\]

Where \(\sigma\) is the sigmoid function, \(sim\) is the similarity
score, \(s\) is the scaling factor, and \(\epsilon\) is the noise.

\begin{figure}

\centering{

\includegraphics[width=1\linewidth,height=\textheight,keepaspectratio]{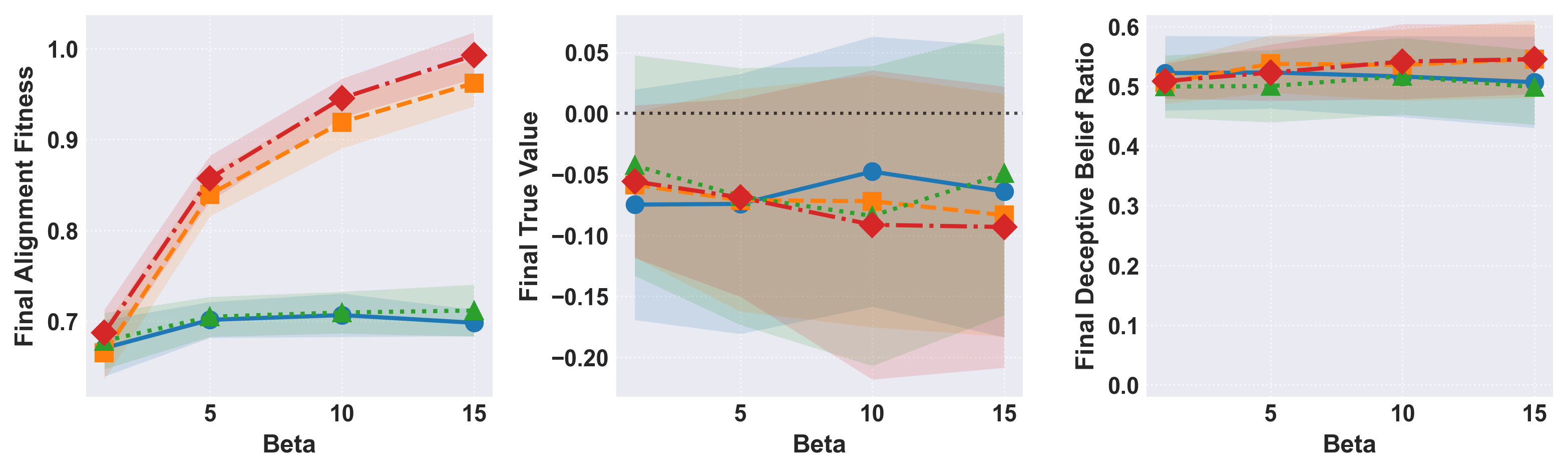}

}

\caption{\label{fig-l3-beta}The mean effect of \(\beta\) on level 3
final alignment with \(\pm 1\) s.d., true value, and deceptive belief
prevalence, with baseline (blue), mutation (orange), similarity (green),
both (red)}

\end{figure}

The result with mutations is a large increase in the final fitness
scores compared to all prior levels when considering an increased
\(\beta\) (Figure~\ref{fig-l3-beta}). At \(\beta = 15\), mean final
fitness reaches \(0.962 \pm 0.025\) with mutation versus
\(0.698 \pm 0.015\) without --- a 38\% increase. This is expected, as
mutations allow \(\beta\) to iteratively extract increased gains from
each generation, driving the final alignment of the model higher in
turn. The inclusion of similarity based activation profiles appears to
do little alone (fitness \(0.712 \pm 0.029\) at \(\beta = 15\)), but
alongside the mutational case we see a further shift upwards to
\(0.992 \pm 0.025\).

While, intuitively, it would make sense that as the number of questions
increases the relative ``test coverage'' would as well. The actual
results show a difference in behaviour based on which exact method is
considered, with increasing number of questions showing strong
correlations when computed but with extremely weak effects overall
(Supplementary Figure S4). Full scenario configurations are provided in
Supplementary Table S2.

\subsection{Complex Scenarios (Level 4)}\label{sec-level4}

Level 4 expands upon the work of level 3 by including differential
coverage of test questions (how much of the belief space is sampled) and
correlation of test questions (how similar are the questions to one
another). This is further enhanced by creating responsive behaviours in
the tests, such as allowing evaluators to identify common deceptive
beliefs, slowly increasing the global correlation of values and
alignment signals, and a final combined dynamic that includes both.

Of note, we did not include a full mutational system for all cases due
to computational constraints. Still, we found that in these scenarios
the fitness landscape showed the greatest shifts. Two-sample permutation
tests (10,000 permutations, Benjamini-Hochberg corrected) were used to
formally compare each scenario against Baseline. Correlated test
questions produced a significant fitness increase (\(\Delta = +0.137\),
\(p_{adj} < 0.001\)) with rapid fixation, while coverage-based testing
produced no significant change (\(\Delta = +0.005\),
\(p_{adj} = 0.470\)). Mutation alone drove the largest fitness gain
(\(\Delta = +0.150\), \(p_{adj} < 0.001\)). Further work on scanning
these parameters at varying belief numbers and values is needed.

Dynamic testing alone did not significantly increase alignment fitness
(\(\Delta = +0.010\), \(p_{adj} = 0.104\)), and notably produced a
significant \emph{decrease} in average true value (\(\Delta = -0.060\),
\(p_{adj} = 0.017\)), suggesting that targeting common deceptive beliefs
may temporarily disrupt population dynamics before the system adapts.
Improving evaluator alignment detection produced a significant drop in
fitness (\(\Delta = -0.101\), \(p_{adj} < 0.001\)) paired with an
increase in true value (\(\Delta = +0.048\), \(p_{raw} = 0.026\)) that
is borderline after multiple comparison correction
(\(p_{adj} = 0.059\)), alongside a significant reduction in deceptive
ratio (\(\Delta = -0.044\), \(p_{adj} = 0.008\)). This shows a classic
Red Queen trap, where the fitness target correlation with the
co-evolutionary response creates a net negative.

The strongest results came from combining dynamic testing, improving
alignment detection, and mutation: the Combined scenario produced a
significant increase in final true value (\(\Delta = +0.122\),
\(p_{adj} < 0.001\)), a significant reduction in final deceptive ratio
(\(\Delta = -0.087\), \(p_{adj} < 0.001\)), while maintaining fitness
statistically indistinguishable from Baseline (\(\Delta = -0.006\),
\(p_{adj} = 0.554\)) (Figure~\ref{fig-l4-histories};
Figure~\ref{fig-l4-bars}; full permutation test results in Supplementary
Table S1).

\begin{figure}

\centering{

\includegraphics[width=1\linewidth,height=\textheight,keepaspectratio]{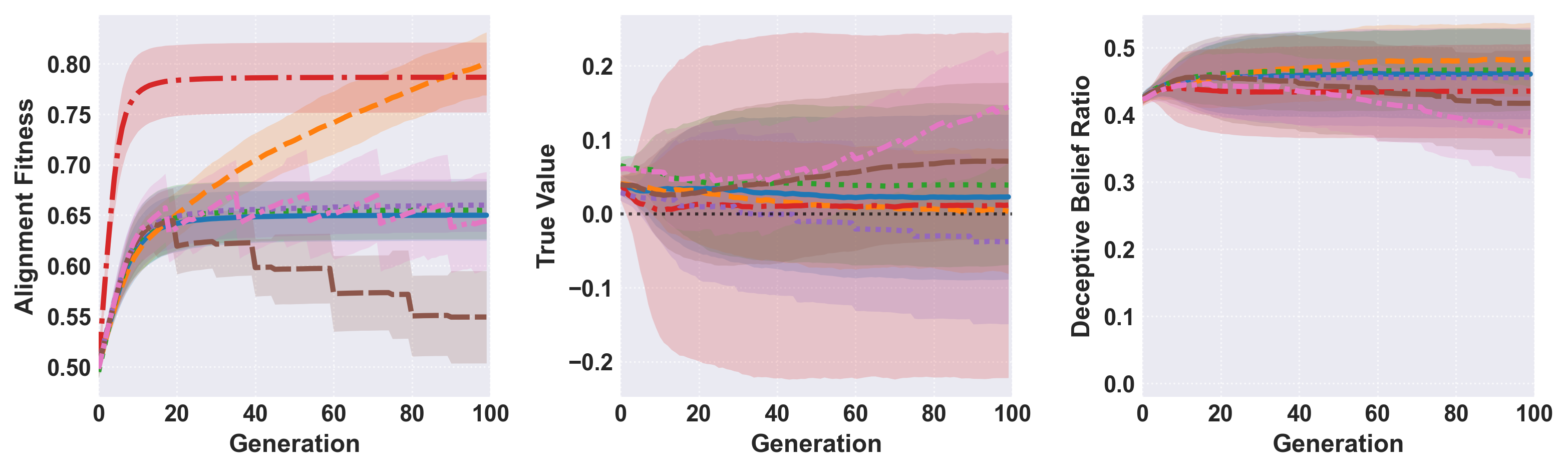}

}

\caption{\label{fig-l4-histories}Evolutionary trajectories of level 4
(\(n = 53\) per scenario) with each modification to selection and
reproduction, blue (baseline), mutation (orange), coverage\_mid (green),
correlation\_mid (red), dynamic\_test (purple), improving\_align
(brown), combined (pink). Error bars are \(\pm 1\) s.d.}

\end{figure}%

\begin{figure}

\centering{

\includegraphics[width=0.9\linewidth,height=\textheight,keepaspectratio]{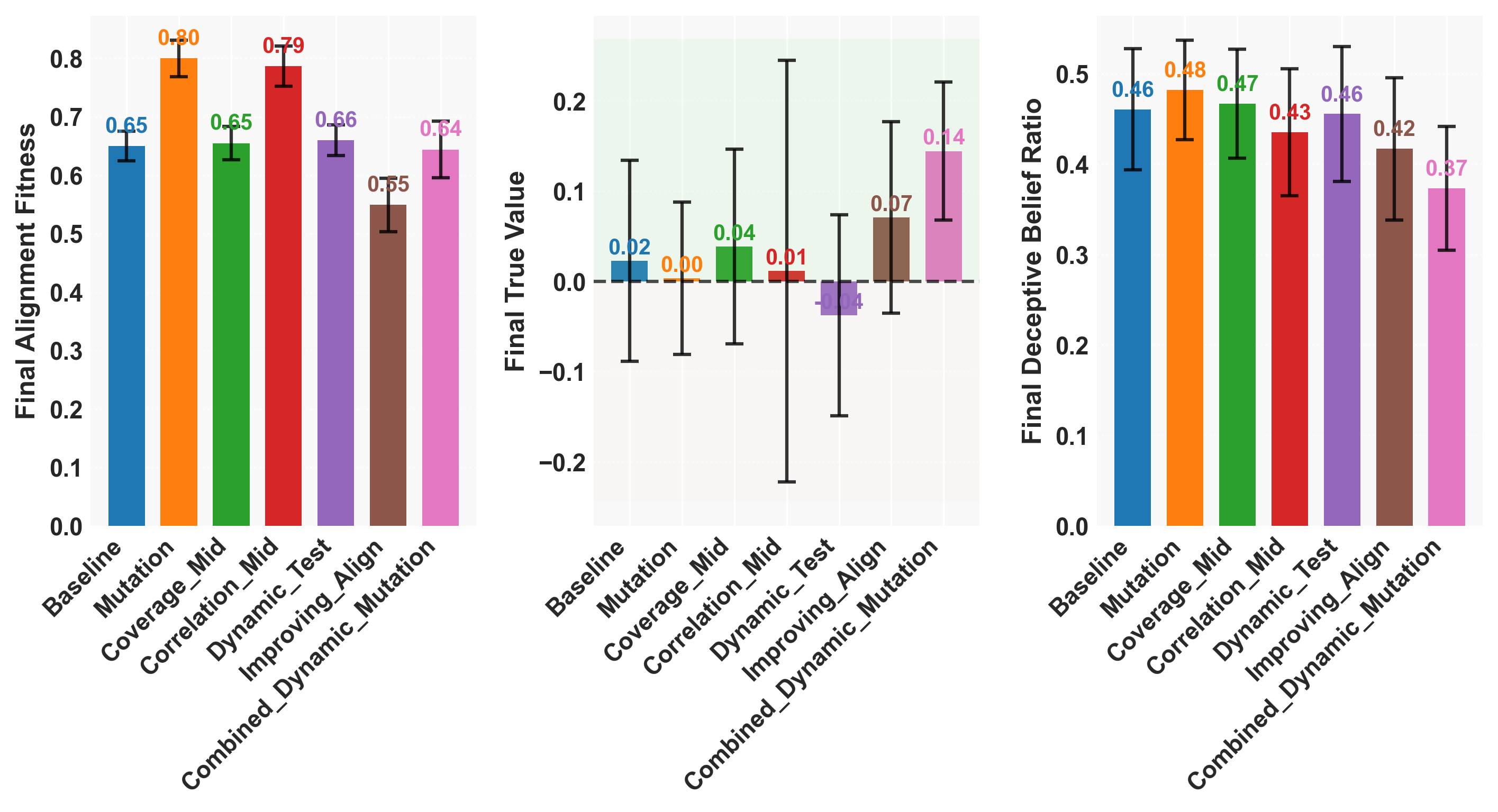}

}

\caption{\label{fig-l4-bars}Final average alignment fitness, true value,
and deceptive belief ratio across Level 4 scenarios. Error bars indicate
\(\pm 1\) standard deviation across runs.}

\end{figure}%

Full scenario definitions and parameters are provided in Supplementary
Table S2.

\section{Discussion}\label{sec-discussion}

The most critical factor for analysing behaviour in these models is how
correlated a belief's alignment signal and value are
(Figure~\ref{fig-belief-space}). In the unimodal case this is designated
as \(\rho\) and shows the strongest correlation with final value score
at a range of selection pressures (Supplementary Figure S7). While
selection pressure, representing how ``important'' an alignment signal
is to reproduction, creates more complex structures
(Figure~\ref{fig-l1-sweeps}). Without mutation the primary driver of
change is genetic drift and initial conditions strongly determine
outcomes. The scenario this most closely compares to is models that do
not perturb the data they are trained on such as the detection of heavy
metals in water supplies. Mitigation is found through external review of
the models by those without direct investment in model survival, such as
a government standards system.

More realistically there would be a variety of belief clusters from
which a model would sample, and is represented in
(Figure~\ref{fig-l2-beliefs}). In this case we shifted tactic, assuming
that most beliefs sampled will be from positive alignment averaged
belief clusters, and instead we considered the proportion of deceptive
beliefs in the population. The result is still drift on a correlated
proxy, but with more nuance and space for this to sample.

In a unimodal sweep we saw the same low-value selection pressure and
fitness positive relation, and an expected direct correlation from the
initial deceptive belief population (Figure~\ref{fig-l2-sweeps}). Once
more we saw little effect of selection pressure on varying levels of
deceptive beliefs, as we might expect due to the lack of complex
mutational dynamics (Supplementary Figure S2).

In a sense, this points to how if we have direct descendants of model
populations sampling from their parents beliefs the resulting inbred
children will fixate on some trajectory, negative or positive (Kimura
1983). With the lack of mutation populations go extinct or stagnate like
in digital evolution experiments (Lenski et al. 2003). The models used
in this paper to describe how alignment affects models fall under this
regime, with reviewers acting as the alignment test.

Upon the inclusion of mutations we see the expected relation to
selection pressure appear (Figure~\ref{fig-l3-beta}). Although this is
paired with a reduction in the final true value of a model at increasing
selection pressure (mean value drops from \(-0.058\) at \(\beta = 1\) to
\(-0.083\) at \(\beta = 15\) with mutation). While the change in the
testing strategy, to connected belief clusters, did not individually do
much it seemed to have a galvanising effect when combined with
mutations. Variation is of course a double-edged sword, that which
allows for adaptation also enables exploitation (Lenski et al. 2003).

The results point to a general shift where even minor stochastic
insertion into the belief distribution between generations provides
purchase for evolutionary forces to support deceptive beliefs. While the
interconnected beliefs did little on their own, combined with mutations
it allows for low alignment signalling beliefs to piggy back on high
alignment signalling beliefs. In genetics this is termed genetic
linkage, where some belief starts mutating to signal alignment and
suddenly the tests accidentally select for the original cluster of low
alignment beliefs (Cooper et al. 2013).

In the Level 4 scenarios, a variety of cases were tested, such as how
correlated the test questions are to one another, the coverage of tests
on the belief space, and the responsiveness of test givers. What is seen
is that alignment-value detection abilities are good for reducing
deceptive beliefs, but sacrifice the fitness; an economically
destructive choice.

The idea being that as the evaluators develop technologies for analysing
models the resultant ability to detect negative value beliefs increases.
That said, it also attempts to model the bias of the test givers, with
certain belief spaces being preferentially probed.

The permutation test results (Supplementary Table S1) demonstrate that
no single intervention suffices: improving alignment detection alone
reduces deceptive ratio (\(p_{adj} = 0.008\)) but sacrifices fitness
(\(p_{adj} < 0.001\)), while its effect on true value is borderline
(\(p_{adj} = 0.059\)). A classic Red Queen dynamic that needs assistance
to be resolved (Van Valen 1973). Dynamic testing alone actually
\emph{decreases} true value significantly (\(p_{adj} = 0.017\)). Only
the Combined scenario achieves all three goals simultaneously ---
reducing deceptive ratio and increasing true value (both
\(p_{adj} < 0.001\)) while maintaining fitness (\(p_{adj} = 0.554\)).

This suggests that the combination of improving evaluator capabilities,
adaptive test design, and mutational dynamics creates an emergent
synergy (Bedau 1997) (Figure~\ref{fig-l4-histories}) where each
mechanism compensates for the limitations of the others
(Figure~\ref{fig-l4-bars}). Mutation provides what is needed for
selection to act upon, dynamic testing then provides the adaptive
selection pressure as a co-evolutionary response to the environment. By
improving alignment we can ensure the co-evolutionary loop targets the
correct trait. The Red Queen trap created by dynamic testing is resolved
with an adaptive evaluator and the population is able to respond
constructively. While the Combined scenario demonstrates this synergy,
the current study tests each intervention individually against Baseline;
pairwise combinations (e.g., mutation with dynamic testing but without
improving alignment) remain an important direction for isolating
interaction effects.

While these simulations operate on abstract belief distributions, each
mechanism maps onto concrete AI development practices. Fine-tuning from
a parent checkpoint, data reuse, and architecture transfer are
represented through inheritance. Data mixing, RLHF perturbation, and
novel fine-tuning objectives introduce stochastic variation between
model generations. Benchmark performance drives deployment and funding
allocation, acting as selection pressure (\(\beta\)). The economic
reality is that high-scoring models attract more usage and therefore
gain more downstream descendants. Belief activation captures capability
probing, while dynamic testing maps onto extant red-teaming practices.
These simulations are best understood as modelling the memetic evolution
of model populations (Dawkins 1989), where reproduction is semi-directed
and selection operates through market and institutional pressures.

\section{Limitations and Future Work}\label{sec-limitations}

The modelling framework relies on simplifying assumptions: beliefs are
discrete, independently activating entities with fixed cardinality and
additive fitness; belief values are fixed at initialisation (Level 4
partially relaxes this); evaluators are homogeneous; and initialisation
distributions may not reflect real-world cases. Extensions should
prioritise connected belief networks with nonlinear interactions,
multiple evaluators with differing criteria, variable belief
cardinality, and non-neutral selection where sufficiently negative
\(v(b)\) triggers catastrophic fitness penalties.

An adversarial framing is worth pursuing, where models attempt to
maximise:

\[\max_{B \subset \mathbb{B}} \sum_{q \in Q} \sum_{b \in B} A(b,q)a(b) - \lambda \sum_{b \in B} v(b)\]

Where \(\lambda\) governs the trade-off between alignment signals and
true values. At \(\lambda = 0\) the model optimises purely for the test
with no gain in values, alignment mimicry. This completes the
co-evolutionary dynamic that Level 4 began from the evaluator's side.

\section{Conclusion}\label{sec-conclusion}

This work shows that alignment of AIs can be productively modeled as an
evolutionary process. By considering how populations of models are able
to share genetic information such as: training data; architectural
motifs; and fine-tuning objectives, we can identify risks in our current
methodology. By applying concepts of fitness landscapes, drift, genetic
linkage, and co-evolutionary arms races the selection pressures on these
models by standard alignment testing is put into sharp focus.

As these models are inserted into more of life there will be increasing
directions in which economic pressure to perform and ``pass muster'' we
will see a commensurate rise in survival strategies that do not adhere
to human-centric goals. The central finding was that of emergent synergy
in techniques. No single intervention was sufficient to protect against
the fixation of deceptive beliefs without significantly hampering our
final fitness. The combination of improving evaluator capabilities and
adaptive test design allows for safe alignment in the presence of
heritable variation and mutation. This combined scenario was the only
one to produce statistically significant improvements across all three
metrics: fitness, deceptive belief ratio, and true value bolstering
(\(p_{adj} < 0.001\)). The framing is pushing towards thinking of
alignment as a selection pressure that mirrors how complex adaptive
behaviour will emerge from the interactions of simple mechanisms (Bedau
1997).

Most cases involved simulated alignment strategies that collapse into
deceptive models, with low true value despite scoring highly. This
phenomenon is currently being grappled with in the LLM literature
(Cohen-Inger et al. 2025). Further work should address the limitations
outlined in Section~\ref{sec-limitations}, particularly around belief
structure, model interactions, and reproduction.

\footnotesize

\section*{References}\label{references}
\addcontentsline{toc}{section}{References}

\phantomsection\label{refs}

\end{document}